\documentclass[conference]{IEEEtran}
\ifCLASSINFOpdf

\else

\fi

\usepackage{algorithm}
\usepackage{slashbox}
\usepackage{mathrsfs}
\usepackage{amsfonts}
\usepackage{algorithmicx}
\usepackage{algpseudocode}
\usepackage{graphics}
\usepackage{epsfig}
\usepackage{multirow}
\usepackage{amsmath}
\usepackage{amssymb}
\usepackage{textcomp}
\usepackage{booktabs}
\usepackage{subfigure}
\usepackage{multirow}

\begin{document}
%
\title{Heterogeneous Metric Learning with Content-based Regularization for Software Artifact Retrieval}
\author{
\IEEEauthorblockN{Liang Wu\IEEEauthorrefmark{1}\IEEEauthorrefmark{4},
Hui Xiong\IEEEauthorrefmark{2},
Liang Du\IEEEauthorrefmark{3},
Bo Liu\IEEEauthorrefmark{4},
Guandong Xu\IEEEauthorrefmark{5},
Yong Ge\IEEEauthorrefmark{6},
Yanjie Fu\IEEEauthorrefmark{2},
Yuanchun Zhou\IEEEauthorrefmark{1} and
Jianhui Li\IEEEauthorrefmark{1}}
\IEEEauthorblockA{\IEEEauthorrefmark{1}Computer Network Information Center,
Chinese Academy of Sciences, Beijing, China
Email: \{wuliang, zyc, lijh\}@cnic.cn}
\IEEEauthorblockA{\IEEEauthorrefmark{2}Rutgers University, USA,
Email: \{hxiong, yanjie.fu\}@rutgers.edu}
\IEEEauthorblockA{\IEEEauthorrefmark{3}Institute of Software,
Chinese Academy of Sciences, Beijing, China,
Email: duliang@ios.ac.cn}
\IEEEauthorblockA{\IEEEauthorrefmark{4}NEC Laboratories China,
Email: liubo@research.nec.com.cn}
\IEEEauthorblockA{\IEEEauthorrefmark{5}Advanced Analytics Institute,
University of Technology Sydney, Australia,
Email: guandong.xu@uts.edu.au}
\IEEEauthorblockA{\IEEEauthorrefmark{6}University of North Carolina at Charlotte, USA,
Email: yong.ge@uncc.edu}
}


%

\maketitle


\begin{abstract}
The problem of software artifact retrieval has the goal to effectively locate software artifacts, such as a piece of source code, in a large code repository. This problem has been traditionally addressed through the textual query. In other words, information retrieval techniques will be exploited based on the textual similarity between queries and textual representation of software artifacts, which is generated by collecting words from comments, identifiers, and descriptions of programs. However, in addition to these semantic information, there are rich information embedded in source codes themselves. These source codes, if analyzed properly, can be a rich source for enhancing the efforts of software artifact retrieval. To this end, in this paper, we develop a feature extraction method on source codes. Specifically, this method can capture both the inherent information in the source codes and the semantic information hidden in the comments, descriptions, and identifiers of the source codes. Moreover, we design a heterogeneous metric learning approach, which allows to integrate code features and text features into the same latent semantic space. This, in turn, can help to measure the artifact similarity by exploiting the joint power of both code and text features. Finally, extensive experiments on real-world data show that the proposed method can help to improve the performances of software artifact retrieval with a significant margin.
\end{abstract}


%
\IEEEpeerreviewmaketitle

\section{Introduction}
Software artifact retrieval, which is also frequently mentioned as software traceability, is of considerable usefulness for developers, since traceability provides insights into system development and evolution assisting in locating software features, analyzing requirements, managing and reusing legacy systems and etc. It gives essential support in understanding the relationships within and across software requirements, design and implementation \cite{ref43}. It is a fundamental task throughout the software development life-cycle, and is especially important for large-scale and complex software systems.

The task has been viewed as a special case of information retrieval in the area of software engineering. The documents to be retrieved are code programs, and queries are often formulated by developers or automatically taken from requirements and bug reports, e.g., titles of bug reports are used to localize bugs in code programs.

In order to retrieve code artifacts, traditional methods try to generate a textual representation for source code by picking up words from programs and then compare them with queries, i.e., transforming the problem to a text retrieval task. Existing methods only leverage textual similarity between queries and the extracted textual content of code, because the similarity between code and text cannot be computed directly.

Along this stream of research, vector space model and stochastic language model are first adopted \cite{ref1,ref2}. The query and codes are both represented as bag-of-words feature vectors. Subsequently, several other information retrieval models, as well as their variants and ensembles, are also experimented to improve the accuracy of the software artifact retrieval task\cite{ref3,ref4,ref5}. These text-based methods are reasonable for the task, since some words may be used in programs by developers. The problem is that these methods have not fully exploited the semantics and operational information embedded in the program. Although the code is composed of text, simply breaking them into a bag of words loses important information and descriptive features. The adoption of proper semantic representation to measure the similarity between code itself and query motivates the work conducted in this paper.

To this end, in this paper we first extract the relationships between functions and classes, including reference, implementation, inheritance, as features because they indicate the associations between different entities.
In addition, as most source code implements a certain target by assembling several functional code fragments, the code patterns which frequently appear in programs bear more semantic than the rest. We attempt to discover and exploit them to serve as the additional features of the corresponding source code. More specifically, we first extract code relationship features of code programs, and then organize the code as a tree structure, in which each node depicts the adjacent program statements separated by the block delimiters. The tree nodes are then combined iteratively in an agglomerative manner, and frequent code patterns are extracted as code features. Thus the extracted code features contain both the textual content of the code and also preserve the functional information by clustering the logically related statements together. The details are presented in Section \ref{code}.


Although homogeneous distance metric learning \cite{ref40,ref41} has been proposed to learn a good metric to compare two objects and has played a significant role in statistical classification and information retrieval, it requires the objects to share identical features and are comparable. The extracted code features and text features in our task, however, are heterogeneous, so they cannot be compared directly. Thus we propose a novel heterogeneous distance metric learning method to discover the shared semantic representation of text and code features. The proposed approach in our work enables the distance calculation between instances from different feature spaces. More specifically, two transformation matrices are produced to map the code-feature representation of code and the text-feature representation of queries into a shared semantic space, where codes and queries own a homogeneous low-rank representation. Then queries and programs can be compared directly. Similar ideas have been proposed to solve the problem of cross modal multimedia retrieval, such as using text as queries to retrieve music and using text to retrieve pictures \cite{ref7,ref8,ref21,ref22}.
Researchers proposed to build distance metric between different media and help to get better results.
However, since the code features may contain words that overlap with the words in queries, in our task the heterogeneous features are partially comparable. The overlapped words, which may be useful for retrieving software artifacts, will be lost, if we simply use the multimedia retrieval techniques. Thus, we further propose a data matrix to preserve the content-based similarity between text features and code features, which is further adopted to regularize the objective function by constraining the parallel features to be similar in the new space. The content-based regularization exploits the similarity between text and code, and is useful for improving other methods from the domain of heterogeneous multimedia information retrieval according to the experimental results.

By integrating the heterogeneous distance metric learning approach and the content-based regularization into a unified framework, we propose a model named Heterogeneous Metric Learning with Content-based Regularization (HMLCR) to build a distance metric between the feature spaces of code and text.
Extensive experiments are conducted to verify the effectiveness of the proposed method. The experiments are based on real-world data sets, which are obtained from two open-source applications with distinct functions and different programming languages.

The main contributions of the paper lie in four aspects:
\begin{enumerate}

\item  Two kinds of code features, namely code relationship features and frequent code snippet features are proposed to better represent programs, which contain the operational and functional information of codes and help to overcome the bottlenecks of traditional methods in the area of software engineering.

\item  A novel heterogeneous distance metric learning model is proposed to allow similarity computation between codes and text. The similarity is then used to enhance the retrieval results.

\item  We evaluate our method using real world open source software data. The experimental results demonstrate that the proposed model can improve the prediction accuracy significantly and outperform several baselines.

\item We demonstrate a case study using HMLCR to retrieve corresponding word features for code features, which are used to further explain the changes brought by incorporating the proposed model.
\end{enumerate}

The remaining of the paper is organized as follows. The problem definition of software traceability and the proposed approach are presented in Section \ref{approach}. Experiments are conducted in Section \ref{experiment}, where the experimental results and a case study are provided. Then we discuss close related topics in Section \ref{relatedwork} and Section \ref{conclusion} concludes the work and provides possible future directions.

\begin{figure*}
 \centering
  \subfigure[Training of the proposed method. Two transformation matrices are generated.]
  {
    \label{structure1} 
    \includegraphics[width=0.470\textwidth]{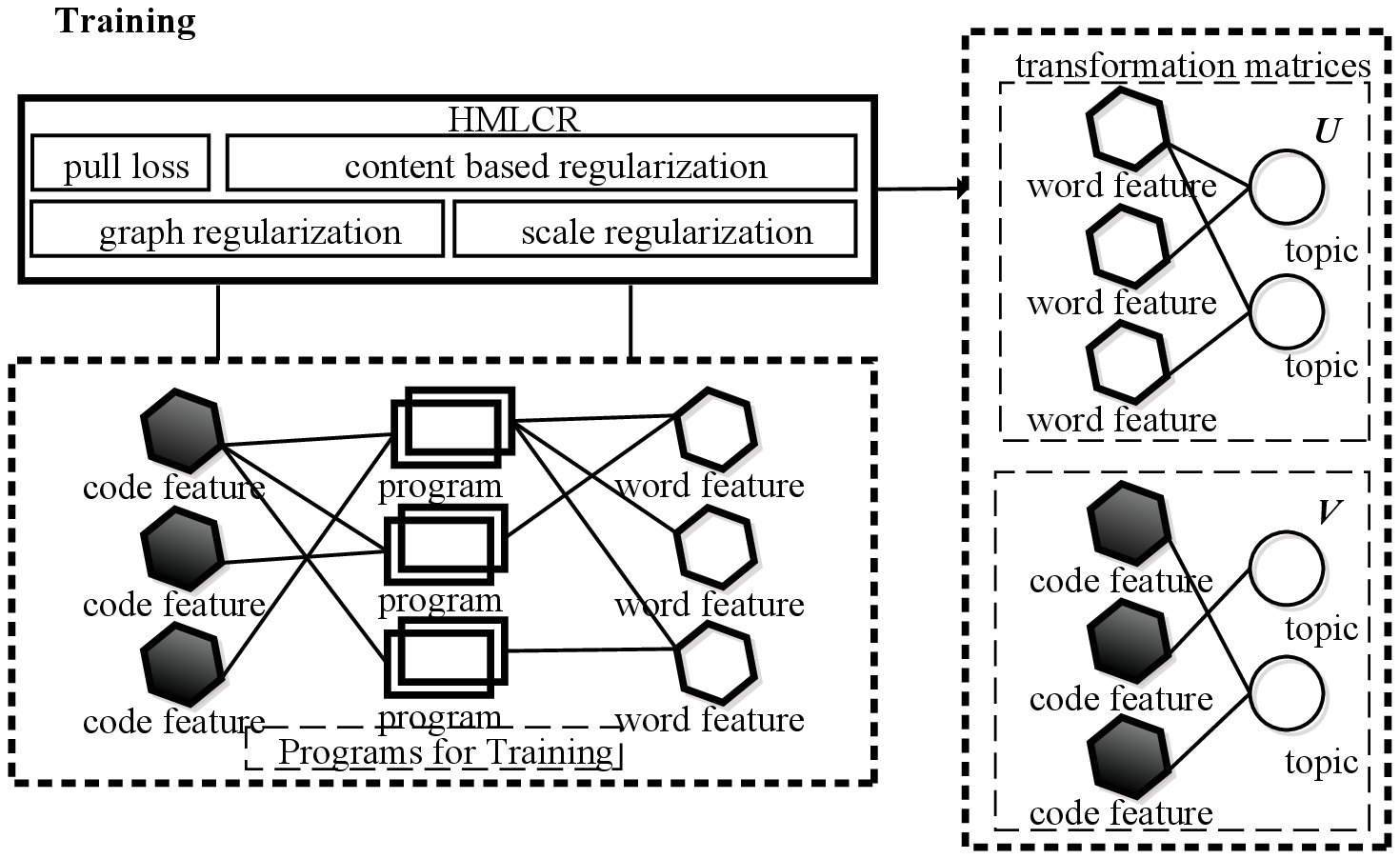}
  }
  \subfigure[Testing of the proposed method. Text-code similarity is calculated based on the two transformation matrices and text-text similarity is calculated directly.]
  {
    \label{structure2} 
    \includegraphics[width=0.470\textwidth]{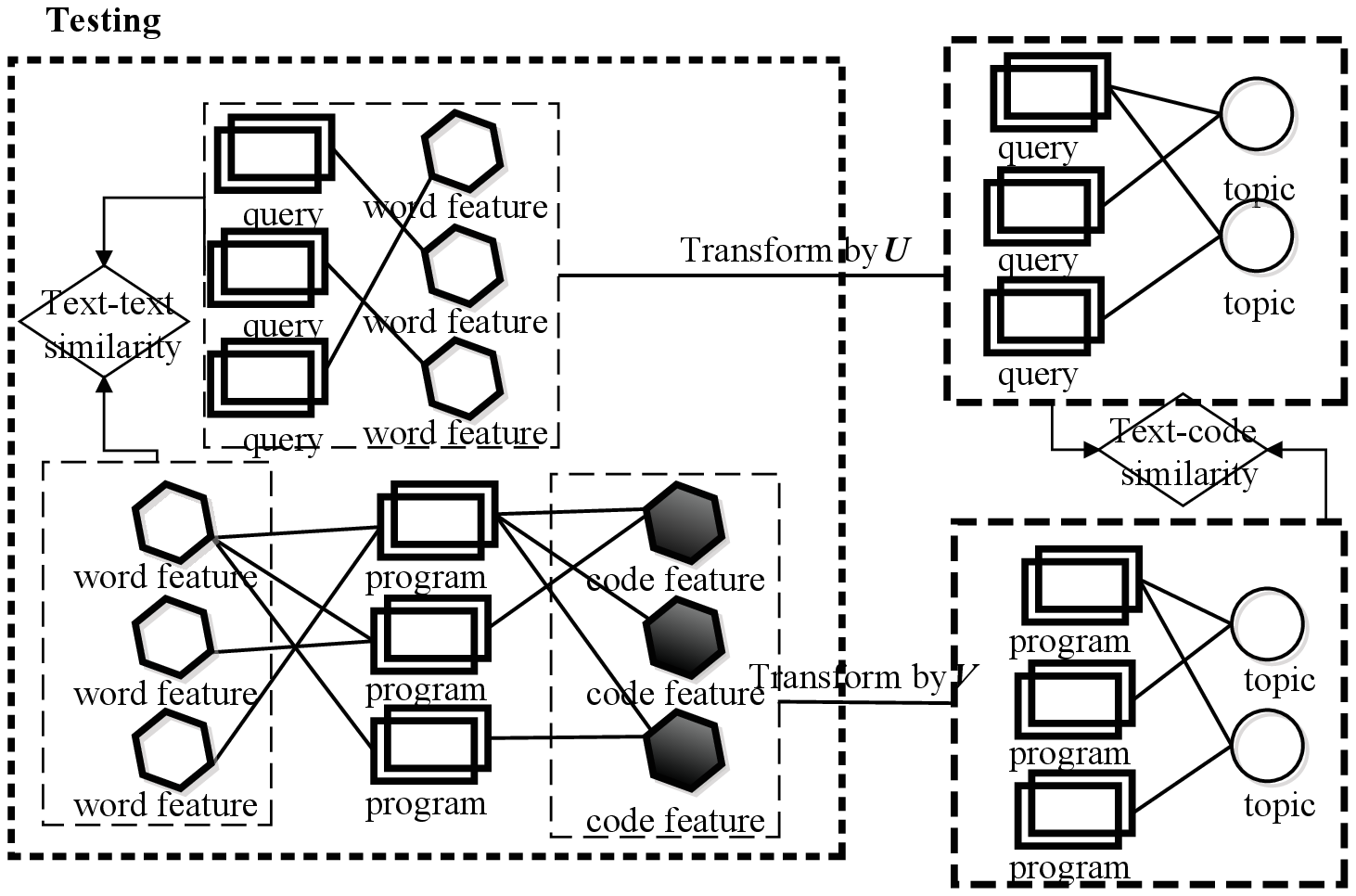}
 }
  \caption{The framework of our proposed approach.}
  \label{newframe} 
\end{figure*}
\section{Proposed Approach}\label{approach}
As the code contains useful clues for linking text, code and between them, our target is to directly compute the similarity between code and queries, and further use the similarity to enhance the retrieval results. To allow computation in different feature spaces for code and text, two transformation matrices are constructed to map the text queries and the code files into a same semantic space. Thus, in this section, we first introduce the definition of the problem and then describe the framework we proposed. Two kinds of code features are discussed subsequently. Finally, the HMLCR model and its optimization are presented.
\subsection{Problem Definition}\label{definition}
The task has been viewed as a general information retrieval task, where the queries are written in natural languages, and the code has two parts, containing both source code and textual information consisting of comments, identifier names and etc. Here we denote the code as $\mathbb{D}=\{(x_{1}^{c},x_{1}^{d},l_{1}),\cdots,(x_{m}^{c},x_{m}^{d},l_{m})\}$, which contains $m$ code programs, each consisting of the source code $x^{c}$ and textual information $x^{d}$. Every code has a label $l_{i}$, which represents the specific function of the code. Labels are generated either manually or according to certain disciplines of a specific domain. For most projects, every code file has a unique identifier. But in some real applications, the label is used to denote the function of a code file. Thus different programs may share an identical label. The query set is denoted as $\mathbb{Q}=\{(x_{1}^{d},l_{1}),\cdots,(x_{n}^{d},l_{n})\}$, where each query contains a short paragraph describing itself and a label which is taken from the same vocabulary as the labels of code. The source of queries includes bug reports, specifications and requirements, which normally contain references to code files that a query describes. The reference is used to produce labels of queries.
\newtheorem{definition}{Definition}
\begin{definition}\quad (Software Artifact Retrieval) Given a program data set $\mathbb{D}=\{(x_{1}^{c},x_{1}^{d},l_{1}),\cdots,(x_{m}^{c},x_{m}^{d},l_{m})\}$ and an unlabeled query, the basic objective of the software artifact retrieval task is to retrieve relevant code files in the unlabeled data set $\mathbb{T}=\{(x_{1}^{c},x_{1}^{d}),\cdots,(x_{n}^{c},x_{n}^{d})\}$.
\end{definition}

Traditional methods focus on computing the similarity between queries and the textual representation $x^{d}$ of code files. The contribution of our work is to incorporate the similarity between queries and $x^{c}$. In order to enable the similarity computation, we extract code features from $x^{c}$, and build heterogeneous distance metrics between code features and word features.
\subsection{Overview}\label{overview}
Figure \ref{newframe} presents the framework of our proposed approach. The word features are first extracted from textual content of codes. Since words cannot fully reveal functional information of program codes, we further propose code features to overcome the bottleneck. The code features are obtained by extracting the code relationships and iteratively combining the adjacent program expressions based on their structure. Since code and word features are heterogeneous, in the training process, as depicted in Figure \ref{structure1}, two transformation matrices $U$ and $V$ are built by the HMLCR model. The two transformation matrices can be viewed as the heterogeneous distance metric between the text features and the code features, and are used to enable the direct comparison of similarity between heterogeneous feature spaces. Although it is hard to characterize the intrinsic structure of the homogeneous and heterogenous data in the new space, fortunately, various contextual information is available in real applications of software artifact retrieval, which can be leveraged to assist the learning process. Here, we put forward four constraints on the generation of the distance metrics: 1)$pull\ loss$ is used to force the textual representation and code representation of programs with the same label to be near in the new space. 2)$graph\ regularization$ is employed to gather both homogeneous and heterogeneous entities with a same label together in the new space. 3)As some code features contain words, and these features should be similar to the word features which also contain the words in the new space. Thus, the $content\ based\ regularization$ aims to reserve this similarity by constraining similar features to be related in the new space after the projection. 4)A $scale\ regularization$ item is adopted to avoid over-fitting. The HMLCR model focuses more on the relationships between textual representation and code representation of programs, as in real world applications, they are born together and thus the links are sufficient for training; while the query-code relationship is not taken into consideration, since the links between queries and code are much harder to collect and far less than enough.

As depicted in Figure \ref{structure2}, the final similarity between a query and a code file is two-fold: text-text similarity can be directly computed between the textual representation of code programs and queries, as they are homogeneous; text-code similarity is acquired by transforming the queries and code into a common feature space and representing them by hidden topics. The linear ensemble of them is taken as the final similarity.
\subsection{Feature Extraction for Code Programs.}\label{code}
Most existing work on source code retrieval focuses on extracting the words in programs, that is, collecting words to build documents from comments, identifier types, function names and etc., which ignores the structures of the programs. This line of methods may encounter difficulties when the textual content is not sufficient. For example, As the queries and the code are often written by different organizations, a variety of vocabularies may be employed to characterize a similar or even same entity/behavior. This results in a gap between the queries and the code descriptions.

We observe that, though different words are used to describe a same thing, the specific lines of programs, which implement the corresponding behavior or entity, almost keep still from time to time. Based on the observation, a natural intuition is that the specific lines of code may be extracted and leveraged to bridge the gap between different queries and programs during retrieval.

The code features are extracted as the bridges. They are used to gain more information by linking them to the frequently co-occurred words. Since the content of code is also useful for representing programs, we also adopt the content of code features during optimization.
\subsubsection{Code Relationship Feature Extraction}
The first kind of code features are the code relationship features. Between different code files, some linking information is explicitly announced. The related information between different code files indicate the associations between different entities, and thus are useful for inferring the semantic of code files. So we extract the relationships between functions and classes, including reference, implementation, inheritance, as code relationship features.
\subsubsection{Code Snippet Feature Extraction}
A key issue of extracting frequent code patterns is how to split a whole program file into pieces. As most source code can be logically formalized like a tree structure, i.e., the whole program is the root node and its sub modules are first layer children. We collect these nodes to build the candidate set of our code features. Each node represents the expressions which are separated and combined by the block delimiters, since expressions of source code are often separated to different domains based on the ending characters, like the curly brackets ``\{'' and ``\}'' in C and Java. The adjacent code expressions are then merged together as a tree node and then added to the code tree. They are further merged into their higher level parents. Therefore, we extract the nodes hierarchically, from leaf nodes to the root node.

Algorithm \ref{codefeaturecandidate} displays the algorithm of the extraction and vectorization of code snippet features. The generated tree nodes of programs are sequentially checked by the algorithm. The mapping between code and features are saved to produce the data matrix, and the candidate set $C$ reserves all code nodes. We name the extracted code nodes as code snippets. Here, a code snippet consists of at least one expression of source code. As stated earlier, we first merge the program statements between the delimiters as the lowest level snippets; and we then merge the snippets together according to the positions of the delimiters, as the snippets are normally organized and nested to perform more complex behaviors. These snippets are then combined in a agglomerative manner according to the hierarchical organization of the program, thus a set of code snippet features is generated.
\begin{algorithm}[htb]
\renewcommand{\algorithmicrequire}{\textbf{Input:}}
\renewcommand{\algorithmicensure}{\textbf{Output:}}
\caption{Construction of Code Snippet Feature Candidate Set}
\label{codefeaturecandidate}
\begin{algorithmic}[1]
\Require
The dataset of code programs: $C$.\
\Ensure
The candidate set of code features: $F$;
The map between program files and code features: $M$.
\For{each program $c$ in $C$}
    \State Extract All Nodes $N$ from $c$
    \For{each node $n$ in $N$}
        \If{$n\notin F$}
            \State Add $n$ to $F$
        \EndIf
        \State Add the Program Feature Map to $M$
    \EndFor
\EndFor
\\
\Return $F$ and $M$
\end{algorithmic}
\end{algorithm}
It should be noted that, as we call the set of code snippets as code feature candidates, not all code snippets are finally used afterwards. A lower and a higher bound of number of occurrences are set to filter out those rare and common snippets. The rare snippets, which seldom appear in programs, make the relationship between features and codes sparser; the code feature candidates which appear too frequently in the dataset are less informative for the retrieval task. In this paper, we set the bounds empirically.

Our basic intuition to extract code snippets is they can be used to accurately represent a behavior or an entity, and are useful to bridge the different text and enrich the sparse textual information, since code is function-dependent, i.e., code snippets with same functions should share a same structure. In order to avoid taking in too much noise, a set of preprocessing operations, like transforming the identifer names to identifer types, are performed on the programs before features being extracted. A similar line of methods that are also designed for this goal of removing the useless expressions has been proposed and studied well in the area of software engineering, which is formally named as program slicing method \cite{ref9}.

After extracting code features, including code relationships and code snippets, our objective is to infer its semantic by mining the relationships between code and text, thus to allow direct comparison between code and text.
\subsubsection{Content Information Extraction}
As discussed in Section \ref{overview}, the content of code features is useful for the distance metric learning. Thus, we propose to use a data matrix $R$ to save this content-based similarity between heterogeneous features. As introduced in Table \ref{notations}, $R$ is a $d^{x}\times d^{y}$ matrix, in which each entry $r_{i,j}$ represents the similarity between text feature $i$ and code snippet $j$.

For simplicity, the similarity is defined to be 1 if the text feature $i$ and the code feature $j$ contain a same word. A problem is that, in code snippet features, as the high level tree nodes contain many low level nodes, they may be similar to too many text features erroneously. In order to avoid this, the words are encoded in the first layer and no longer available when they are wrapped more than once. That is, in a code feature, the word is only available in its lowest code snippet. When the code snippet is iteratively combined in the higher code snippet, the word is wrapped in its lowest feature and will not be included as code content more than one time.

\subsection{Heterogeneous Metric Learning with Content-based Regularization}\label{proposedapproach}
\begin{table}
\normalsize
\centering
\caption{Notations and corresponding descriptions}
\begin{tabular}{ll}
\hline
m&number of training code files and queries \\
$d^{x}$&dimensionality of textual feature space  \\
$d^{y}$&dimensionality of code feature space \\
$\lambda_{1\cdots3}$&regularization parameters\\
$\mathbf{X}$&$d^{x}\times m$ matrix of words extracted from code \\
$\mathbf{Y}$&$d^{y}\times m$ matrix of code features\\
$\mathbf{U}$&$d^{x}\times k$ transformation matrix text features\\
$\mathbf{V}$&$d^{y}\times k$ transformation matrix code features\\
$\mathbf{R}$&$d^{x}\times d^{y}$ matrix of similarity between features\\
\hline
\end{tabular}
\label{notations}
\end{table}

Distance metric learning \cite{ref23} has attracted much attention in the last decade. The objective of distance metric learning is to learn a linear transformation matrix to map the data into a new space. Then the distance between objects in the new space is defined by the distance metric. The learning process forces the new space to preserve the similarity and dissimilarity between any two objects in the training data. Distance metric learning proves to be useful in many machine learning tasks, as it helps to capture the optimal distance between two objects.

In the area of cross modal multimedia retrieval, researchers proposed heterogeneous distance metric learning methods to compare different media \cite{ref6,ref7,ref8,ref10,ref19,ref20,ref21,ref22}. Unlike the homogeneous distance metric learning, heterogeneous distance metric learning aims to learn multiple transformation matrices. Each matrix is used to transform the corresponding kind of media to the semantic space.

In Section \ref{code}, we have described the features extracted from code. In order to calculate the similarity between code programs and textual descriptions, we regard code semantic representation as a new media and propose the HMLCR model, to learn two transformation matrices $U$ and $V$ for text and code via optimization.

After that, each textual representation and code feature representation are projected to a new space with dimensionality of $k$. $k$ can be viewed as the number of latent topics. For simplicity, $k$ is set empirically in this work and is normally smaller than $dx$ and $dy$. Then based on the new space, the heterogeneous similarity can be measured. Later in this section, the details of HMLCR are introduced.

We first describe our loss function $\epsilon_{pull}$ that helps to build up the connections between code and text feature spaces. Then we introduce a joint graph regularization item $g$ which constrains the new space to reserve the links between the textual and code feature-based representation of programs. Then we show how to leverage the content information of code features to regularize the training objective. The optimization objective and the optimization method are presented in the end. Some of the notations are summarized in Table \ref{notations}.

\subsubsection{Loss Function}
In order to reserve the relationships between code feature representation and textual representation, we propose to use a loss function to build the new space. As depicted in Equation \ref{loss1}, the item $\epsilon_{pull}(U,V)$ penalizes the large distance between code and its corresponding textual representation, i.e., pulling them together.
\begin{equation}\label{loss1}
\epsilon_{pull}(U,V)=\frac{1}{2}||X^{T}U-Y^{T}V||_{F}^{2}
\end{equation}
\subsubsection{Joint Graph Regularization}
The similarity between code feature and code feature based representation, the similarity between textual and textual representation are also useful for heterogeneous metric learning. Thus, in order to force the low dimensional representations to reserve the relationships, we introduce a joint graph regularization term
\begin{equation}\label{laplacian}
g(U,V) = \frac{1}{2}tr(O\bar{L}^{T}O^{T})
\end{equation}
where $O$ is a $c\times(m+m)$ data matrix which represents the coordinates of the original data in the new semantic space:
\begin{equation}
O =\left(
   \begin{aligned}
U^{T}X,V^{T}Y \\
   \end{aligned}
   \right)
\end{equation}
and $L$ is the normalized graph Laplacian of $W$, which is produced as follows:
\begin{equation}
\bar{L} = I - D^{-\frac{1}{2}}WD^{-\frac{1}{2}}
\end{equation}
in which $W = {w_{ij}}_{(m+m)\times(m+m)}$ means the relationship between the $i$-th object and $j$-th object. The symmetric similarity of labels is encoded into the matrix as follows:
\begin{equation}
w_{ij} = \left\{
             \begin{array}{ll}
             {1,}& {l_{i}=l_{j} \wedge i \neq j;} \\
             {0,}& {otherwise} \\
             \end{array}
        \right.
\end{equation}
Here, $I$ is an identity matrix and D is a diagonal matrix with each entry $d_{ii}$ equals the sum of the corresponding row of $W$, namely $\sum_{j}^{m+m}w_{ij}$.

As denoted in Equation \ref{fourparts}, the matrix $\bar{L}$ can be decomposed into four parts:
\begin{equation}\label{fourparts}
\bar{L} =\left(
   \begin{aligned}
\bar{L}^{xx} \bar{L}^{xy} \\
\bar{L}^{yx} \bar{L}^{yy} \\
   \end{aligned}
   \right).
\end{equation}
thus Equation \ref{laplacian} can be rewritten as:
\begin{eqnarray*}
& &g(U,V)=\frac{1}{2}tr(O\bar{L}O^{T})\\
&=&\frac{1}{2}tr(U^{T}X\bar{L}^{xx}X^{T}U)+\frac{1}{2}tr(U^{T}X\bar{L}^{xy}Y^{T}V)\\ &+&\frac{1}{2}tr(V^{T}Y\bar{L}^{yx}X^{T}U)+\frac{1}{2}tr(V^{T}Y\bar{L}^{yy}Y^{T}V)\\
\end{eqnarray*}
where $tr(U^{T}X\bar{L}^{xx}X^{T}U)$ is the trace of the matrix $U^{T}X\bar{L}^{xx}X^{T}U$. $g(U,V)$ formulates pairwise similarity between the code feature representation and the textual representation of programs using graph. It improves the smoothness of the mappings by penalizing the functions that change abruptly on the joint data graph \cite{ref13}.
\subsubsection{Content-based Regularization}
As mentioned in Section \ref{code}, $R$ is the content-based similarity between code features and text features. To force the transformation matrices $U$ and $V$ to reserve it, we incorporate another term to regularize the objective as follows,
\begin{equation}
c(U,V)=\frac{1}{2}||UV^{T}-R||_{F}^{2}
\end{equation}
where $UV^{T}_{dx\times dy}$ denotes the associations between features in the new space, so $c(U,V)$ aims to penalize the large distance between code features and text features which share the same words.
\subsubsection{Optimization Objective}
Finally, we introduce a regularization term $r(U,V)$ to control the scale of the transformation matrices $U$ and $V$. In this paper, we define the regularization function as follows:
\begin{equation}
r(U,V) = \frac{1}{2}||U||_{F}^{2}+\frac{1}{2}||V||_{F}^{2}
\end{equation}
Thus, the optimization objective is the combination of the aforementioned items:
\begin{equation}\label{objective}
\arg\min\limits_{U,V}\quad \lambda_{1}\epsilon_{pull}+\lambda_{2}g+\lambda_{3}c+r
\end{equation}
where $\lambda_{1}$, $\lambda_{2}$ and $\lambda_{3}$ control the impact of each constraint.
\subsubsection{Optimization}
We will introduce how to minimize the optimization problem in Equation \ref{objective}. The optimization is an unconstrained optimization problem with two matrices $U$ and $V$, which is not jointly convex to them. Since we cannot get the closed-form solutions, we turn to solve the problem by fixing one matrix and optimize the other iteratively. Therefore, gradient descent is adopted to approach the optimal results. By taking the derivatives over the objective function, we have the gradient of $U$ as:
\begin{eqnarray*}\label{gradient1}
\frac{\partial Loss}{\partial U}&=&\lambda_{1}X(X^{T}U-Y^{T}V)\\
&+&\lambda_{2}(X\bar{L}^{x}X^{T}U+X\bar{L}^{xy}Y^{T}V\\
&+&\lambda_{3}(UV^{T}-R)V+U
\end{eqnarray*}
and the gradient of $V$ as:
\begin{eqnarray*}\label{gradient2}
\frac{\partial Loss}{\partial V}&=&\lambda_{1}Y(Y^{T}V-X^{T}U)\\
&+&\lambda_{2}(Y\bar{L}^{yx}X^{T}U+Y\bar{L}^{y}Y^{T}V)\\
&+&\lambda_{3}(UV^{T}-R)^{T}U+V
\end{eqnarray*}
The optimization process can be found in Algorithm \ref{optimizationalgorithm}. Note that the matrices $U$ and $V$ are initialized (line 1) based on the Cross-modal Factor Analysis algorithm\cite{ref10}.
\begin{algorithm}[htb]
\renewcommand{\algorithmicrequire}{\textbf{Input:}}
\renewcommand{\algorithmicensure}{\textbf{Output:}}
\caption{ Iterative Optimization for \textbf{HMLCR}}
\label{optimizationalgorithm}
\begin{algorithmic}[1]
\Require
The data matrices: $X$, $Y$;
The similarity matrices: $W$, $R$;
The parameters, $\lambda_{1\dots 3}$;
The learning rate $\eta$;
The maximal number of iterations $MaxIter$.\
\Ensure
The transformation matrices $U$ and $V$.\
\State Generate U and V\
\For{$i$ = 1 to $MaxIter$}
\State $U$ $\leftarrow$ $U$+$\eta \frac{\partial Loss}{\partial U}$\
\State $V$ $\leftarrow$ $V$+$\eta \frac{\partial Loss}{\partial V}$\
\If{convergence}
\State $break$
\EndIf
\EndFor
\\
\Return $U$ and $V$
\end{algorithmic}
\end{algorithm}

\section{Experiment}\label{experiment}
In this section, we will first introduce the experimental settings as well as the dataset. Then we will describe the evaluation metrics we adopted. Next, we discuss some major results of the experiments. Finally, some case studies are presented and discussed.
\subsection{Dataset}
The model has been incorporated and test in a commercial software to index programs by textual descriptions, but the result cannot be disclosed due to intelligence property issues. So we obtain datasets from two real world open source software, the platform of Eclipse\footnote[1]{http://www.eclipse.org} and Filezilla\footnote[2]{https://filezilla-project.org} and thus make the experimental results reproducible.

Eclipse is a popular open-source IDE for many programming languages written mainly in Java. The project contains approximately 7,000 classes with about 89,000 methods in about 2.4 million lines of code (MLOC).

Filezilla is an open-source FTP client for Windows, Mac OS X and GNU/Linux. The project is written in C and is much smaller than Eclipse, with about 8,012 methods in 410 KLOC.

\begin{figure}[ht]
\includegraphics[height=2.2in]{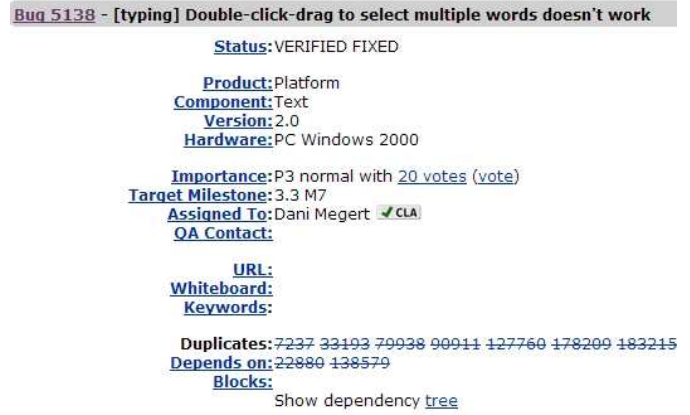}\\
\caption{An example of the online bug report.}
\label{bugreport}
\end{figure}

In order to test our model, we extract titles of bug reports of Eclipse and change logs of Filezilla as queries to retrieve code. The approach of using bug reports and change logs is frequently adopted in the area of software engineering, which is based on change reenactment \cite{ref18}. For example, Eclipse bug 5138\footnote[3]{https://bugs.eclipse.org/bugs/show\_bug.cgi?id=5138}, as depicted in Figure \ref{bugreport}, reports an error on Double Click function of user interfaces. The title ``Double-click-drag to select multiple words doesn't work" is collected as a query and the corresponding code files can be found in the fix patch. We crawled over 1700 bug reports for Eclipse and over 1320 change logs for Filezilla, which own a clear connection to code programs. They are further used for evaluation in our experiments.
\subsection{Evaluation Metrics}
In order to measure the accuracy of the proposed approach, we use three methods to evaluate the retrieval results. The first one is Precision at n(P@n). Here, the position of the true positives is not taken into consideration, as another measure will be adopted to model the usefulness of the returned results in terms of ranking orders. The calculation of P@n is illustrated in Equation \ref{patn}.
\begin{equation}\label{patn}
P@n=\frac{|\{relevant~code~files\} \cap \{retrieved~code~files\}|}{n}
\end{equation}

As in the area of software engineering, traceability module is often designed for professional users who are more patient to click the lower ranked results. Therefore, it is more important to retrieve all the code files given a query, So Recall at n (R@n) is adopted to be the second evaluation metric. Similar with the P@n, the positions of results are not taken to penalize the low ranked right answers. The details of the metric is illustrated in Equation \ref{ratn};
\begin{equation}\label{ratn}
R@n=\frac{|\{relevant~code~files\} \cap \{retrieved~code~ files\}|}{|\{relevant~code~files\}|}
\end{equation}

Another measure method we employ is the normalized Discounted Cumulative Gain(nDCG)\cite{ref12}. This measure is useful for computing the quality of ranking results as it considers both the returned contents and the order of the results. To evaluate the quality of a ranking list, we rank the retrieved code files for each of the queries based on their computed similarity.

In particular, nDCG[p] measures the top $p$ results as follows:

\begin{eqnarray}\label{ndcgcalc}
DCG[p]=rel_{1}+\sum_{i=2}^{p}\frac{rel_{i}}{log_{2}i}\nonumber\\
nDCG[p] = \frac{DCG[p]}{IDCG[p]}
\end{eqnarray}

where $IDCG[p]$ is the $DCG[p]$ value given the ranking list of the ground truth. $rel_{i}$ is the relevance value. Thus, if the higher ranked results are more relevant, a bigger $nDCG$ will be produced, i.e. a higher $nDCG$ represents a better ranking result.

\subsection{Settings}
In order to investigate the effectiveness of the proposed approach, we implemented several methods from the domain of software engineering and multimedia information retrieval as our baselines.
\begin{itemize}
  \item \emph{COS}: The cosine similarity based method \cite{ref1,ref4} which calculates the cosine similarity between the textual representation of code and queries.
  \item \emph{LM}: This method adopts language modeling to calculate the similarity between the textual representation of code and queries \cite{ref1,ref2}.
  \item \emph{LSI}: The latent semantic indexing method that first compresses the textual representation of code and queries and then calculate their similarity \cite{ref3,ref20}.
  \item \emph{CFA}: Cross-modal Factor Analysis model, which is first proposed in \cite{ref10} and is designed to discover the associations between the feature space of different media.
      We adopt this method to calculate the correlations between word features and code features. This method ranks the code files based on two parts, the text-text similarity and the text-code similarity.
  \item \emph{CFA+CR}: In this method, we adopt the content-based constraint to regularize the training process of CFA. Thus, we test the impact of the content based regularization on the retrieval results.
  \item \emph{HMLCR}: The proposed Heterogeneous Metric Learning with Content based Regularization. This method ranks the code files based on two parts, the text-text similarity and the text-code similarity.
\end{itemize}

The first three baselines, COS, LM and LSI, are taken from the area of software engineering, which treat code files as natural languages and extract words from code to form a textual representation.
The other baselines are more similar with the methods in cross media information retrieval, which model the code as heterogeneous media and try to bridge the gaps between different feature spaces.

All these methods can be improved by exploiting application-specific rules, manual labeling and interactive user guidance. In this work, however, we focus on the automatic part of the problem, that is, the similarity computation between code and text. So the baselines we choose, COS, LM and LSI, concentrate more on text comparisons. The rules, labeling and user guidance are not used in our model, either. The proposed method and the baselines can be improved when used in real projects by leveraging these domain knowledge. So the proposed method can be easily equipped to the online systems by replacing the existing calculation modules.

In all experiments, five-fold cross validation is employed. We split the queries into five folds randomly and in each round we select one fold, the queries and the corresponding code files, as the training data, and use the rest for testing. The reported result is the average of all rounds.
\subsection{Experimental Results}
\begin{figure*}
 \centering
  \subfigure[The experimental results based on Eclipse Dataset.]
  {
    \label{parameter1} 
    \includegraphics[width=0.48\textwidth]{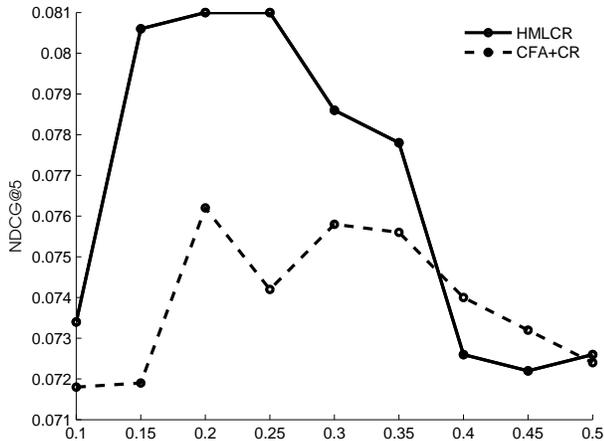}
  }
  \subfigure[The experimental results based on Filezilla Dataset.]
  {
    \label{parameter2} 
    \includegraphics[width=0.48\textwidth]{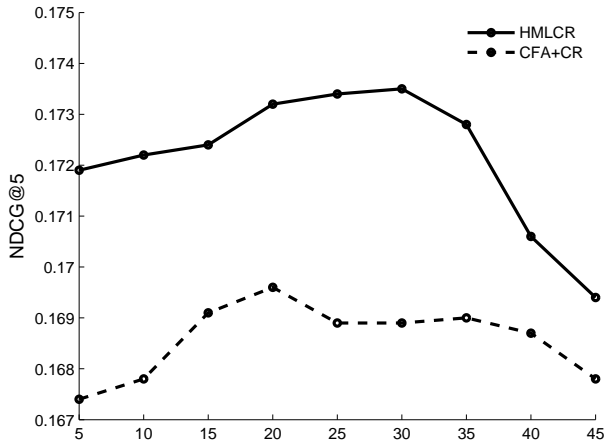}
 }
  \caption{The impact of content-based regularization on different methods.}
  \label{parameter} 
\end{figure*}

\begin{table}[bht!]
\small
\centering
\caption{The precision @ top n results of Eclipse}
\begin{tabular}{l*{4}{c}r}
\hline
        & P@1 & P@2 & P@4 & P@5 \\
\hline
\hline
COS     & 0.016 & 0.0176 & 0.0182 & 0.0196 \\
LM      & 0.0164 & 0.0082 & 0.0123 & 0.0098  \\
LSI     & 0.0092 & 0.0138 & 0.0168 & 0.0159  \\
CFA     & 0.0221 & 0.0246 & 0.0217 & 0.0226  \\
CFA+CR   & 0.0246 & 0.0258 & 0.0242 & 0.0245  \\
HMLCR   & \textbf{0.027} & \textbf{0.0307} & \textbf{0.026} & \textbf{0.026}  \\
\hline
\end{tabular}
\label{resultpatn}
\end{table}

In this section, we describe the retrieval performance of the proposed model and baselines. Table \ref{resultpatn} illustrates the precision at top 1, 2, 4 and 5 for Eclipse dataset. The best result is achieved by the proposed approach and the precision at the first and the second position is obviously improved. An interesting observation is that the CFA+CR method, which incorporates Content-based Regularization (CR) into Cross-modal Factor Analysis model, achieves the second best result. The observation, on the one hand, proves the usefulness of regarding code as multimedia and adopting heterogeneous metric learning to model it; On the other hand, it demonstrates the effectiveness of content-based regularization, as the method significantly outperforms the original CFA method.

\begin{table}[bht!]
\small
\centering
\caption{The precision @ top n results of Filezilla}
\begin{tabular}{l*{4}{c}r}
\hline
        & P@1 & P@2 & P@4 & P@5 \\
\hline
\hline
COS     & 0.0188 & 0.0154 & 0.0191 & 0.0192 \\
LM      & 0.0094 & 0.0094 & 0.0113 & 0.0126  \\
LSI     & 0.0153 & 0.0092 & 0.0105 & 0.0109  \\
CFA     & 0.0751 & 0.0639 & 0.0436 & 0.044  \\
CFA+CR   & 0.1116 & 0.1146 & \textbf{0.0809} & 0.0722  \\
HMLCR   & \textbf{0.1491} & \textbf{0.1156} & 0.0794 & \textbf{0.0748}  \\
\hline
\end{tabular}
\label{resultpatncpp}
\end{table}
Similar findings are observed in Filezilla's P@n results, as depicted in Table \ref{resultpatncpp}. A difference is that the result of CFA+CR baseline draws nearer to that of HMLCR. In the setting of P@4, it even outperforms the proposed approach. The result reveals that the content based regularization has a larger impact on the Filezilla dataset. The phenomena can be explained by the difference of ways of obtaining queries. The queries of Eclipse are extracted from bug reports, which may be filed by ordinary users; while the queries of Filezilla are taken from the change logs, which are issued officially. Thus, the words used in change logs are more refined and accurate, and more closely tending to overlap with the vocabulary of source code.

\begin{table}[bht!]
\small
\centering
\caption{The recall @ top n results of Eclipse}
\begin{tabular}{l*{6}{c}r}
\hline
        & R@1 & R@3 & R@5 & R@20 \\
\hline
COS     & 0.0128 & 0.0348 & 0.0665 & 0.0851 \\
LM      & 0.0164 & 0.0328 & 0.0383 & 0.1052 \\
LSI     & 0.0092 & 0.0275 & 0.0367 & 0.0505 \\
CFA     & 0.0168 & 0.0479 & 0.0719 & 0.1754 \\
CFA+CR  & 0.0176 & 0.0487 & 0.0788 & 0.1778 \\
HMLCR   & \textbf{0.0217} & \textbf{0.0594} & \textbf{0.085} & \textbf{0.1803} \\
\hline
\end{tabular}
\label{resultratn}
\end{table}

\begin{table}[bht!]
\small
\centering
\caption{The recall @ top n results of Filezilla}
\begin{tabular}{l*{6}{c}r}
\hline
        & R@1 & R@3 & R@5 & R@20 \\
\hline
COS     & 0.0069 & 0.0146 & 0.0184 & 0.0259 \\
LM      & 0.0108 & 0.0139 & 0.0457 & 0.2015 \\
LSI     & 0.0113 & 0.0175 & 0.0321 & 0.0796 \\
CFA     & 0.0375 & 0.0776 & 0.1158 & 0.3017 \\
CFA+CR  & 0.0526 & \textbf{0.1435} & \textbf{0.19} & 0.3926 \\
HMLCR   & \textbf{0.0752} & 0.1369 & 0.1883 & \textbf{0.3935} \\
\hline
\end{tabular}
\label{resultratncpp}
\end{table}
The experimental results in terms of recall at top 1, 3, 5 and 20 of Eclipse and Filezilla datasets are displayed in Table \ref{resultratn} and \ref{resultratncpp} respectively. Similar findings can be observed by checking the results. The recall of Filezilla experiment is higher than that of Eclipse on average, since the Filezilla project has much fewer code files and thus the search space is far smaller.

The nDCG results of Eclipse and Filezilla project are shown in Table \ref{resultndcg} and \ref{resultndcgcpp}. As presented in the results, the multimedia methods outperform the baselines from software engineering domain on average, which shows that the functional information of code is more descriptive than the code content. The best result is achieved by considering both the functional and content information: by incorporating the content-based regularization, considerable improvement is encountered on our method and the CFA+CR.

\begin{table}[bht!]
\small
\centering
\caption{The nDCG @ top n results of Eclipse}
\begin{tabular}{l*{5}{c}r}
\hline
nDCG@n    & n=2 & n=4 & n=10 & n=20 \\
\hline
COS     & 0.0256 & 0.0388 & 0.0506 & 0.0526 \\
LM      & 0.0164 & 0.0299 & 0.0415 & 0.0515 \\
LSI     & 0.0183 & 0.0287 & 0.0339 & 0.0339 \\
CFA     & 0.0344 & 0.048 & 0.0734 & 0.0865 \\
CFA+CR  & 0.0356 & 0.052 & 0.0751 & 0.0887 \\
HMLCR   & \textbf{0.0442} & \textbf{0.0589} & \textbf{0.0804} & \textbf{0.0947} \\
\hline
\end{tabular}
\label{resultndcg}
\end{table}

\begin{table}[bht!]
\small
\centering
\caption{The nDCG @ top n results of Filezilla}
\begin{tabular}{l*{5}{c}r}
\hline
nDCG@n    & n=2 & n=4 & n=10 & n=20 \\
\hline
COS     & 0.0171 & 0.022 & 0.0242 & 0.0252 \\
LM      & 0.0153 & 0.0207 & 0.0586 & 0.0765 \\
LSI     & 0.0153 & 0.0226 & 0.0333 & 0.0403 \\
CFA     & 0.0887 & 0.0935 & 0.1482 & 0.1665 \\
CFA+CR  & 0.1491 & \textbf{0.1644} & 0.203 & 0.2347 \\
HMLCR   & \textbf{0.1506} & 0.1598 & \textbf{0.2035} & \textbf{0.2351} \\
\hline
\end{tabular}
\label{resultndcgcpp}
\end{table}
In order to investigate the impact of the content-based regularization in detail, we present the performance of HMLCR and the CFA+CR baseline with a varying $\lambda_{3}$ in Figure \ref{parameter}. In this experiment, Eclipse and Filezilla datasets are used and nDCG is adopted as the evaluation metric, since it can better reveal the quality of the ranking results.

Figure \ref{parameter} presents the results of Eclipse and Filezilla dataset respectively. For Eclipse dataset, as denoted in Figure \ref{parameter1} the best performance is achieved when $\lambda_{3}$ is around 0.15 to 0.30. When the value of $\lambda_{3}$ draws near to 0.0, the performance decreases rapidly to that of the baselines without content-based regularization. When the parameter becomes too large, a reduction is found of the nDCG@5, as it may overwhelm other useful information like the links between text and code files.

Similar observations can be found in the experimental results of Filezilla dataset, as illustrated in Figure \ref{parameter2}. It should be noted that the scale of $\lambda_{3}$ of Filezilla is larger than that of Eclipse. The source of queries of the Filezilla experiment seems to be the causes. As mentioned earlier, the code content regularization is more effective for Filezilla dataset.
\subsection{Case Study}
To further explain the changes brought by the proposed method, HMLCR, several automatically generated cases are displayed in Table \ref{casestudy}. There lies a manually selected case in each row of the table. In each case, a code feature, and the corresponding textual features are given for comparison. The textual features are the top scored words generated by CFA and HMLCR.

There mainly exist two kinds of features, code relationship features and code snippet features in this work. As frequent code patterns are more difficult to understand, we select some class name features from code relationships for simplicity.

The score of each word is computed based on the transformation matrices $U$ and $V$. Since $U$ describes the similarity between words and topics, and $V$ describes the similarity between code features and topics, $UV^{T}$ can be viewed as the similarity between textual and code features in the semantic space.
\begin{equation}\label{relatedness}
R = UV^{T} = \{r_{1} \cdots r_{dy} \}
\end{equation}

As denoted in Equation \ref{relatedness}, the relatedness matrix $R_{dx\times dy}$ consists of $dy$ column vector. Each of the vector represents the relatedness between the code feature and every word feature. The relatedness is adopted as the score in our case study. The top scored words extracted by transformation matrices of CFA are also displayed here for comparison.
\begin{table*}
\small
\centering
\normalsize
\caption{Some selected cases of code features and their corresponding top words scored by HMLCR and CFA}
\begin{tabular}{l*{6}{l}r}
\hline
Feature&Method&\multicolumn{5}{l}{Top scored words}\\
\hline
 \multirow{2}{*}{org.eclipse.swt.swt} & CFA & abstract & file  & parent  & exception  & representation  \\
 & HMLCR & parent  & exception  & format  & representation & abstract \\
\hline
 \multirow{2}{*}{java.io.inputstream} & CFA & cleanup  & file  & immediate & class & overwrite  \\
 & HMLCR  & cleanup  & max  & class & file & firing \\
\hline
 \multirow{2}{*}{org.w3c.dom.css.cssvalue} & CFA & problems & abstract & property & handler & sort  \\
 & HMLCR & problems & handler & property & describe & refresh \\
 \hline
 \multirow{2}{*}{java.io.ioexception} & CFA & file & impact & abstract & describe & warning  \\
 & HMLCR & exception & impact & reconcile & acyclic & abstract \\
\hline
\end{tabular}
\label{casestudy}
\end{table*}

Though some bad cases are still existing in the results of HMLCR, our results look better than the results of CFA obviously. For example, in the first case, ``format'' is more related to the class ``org.eclipse.swt.swt'' than the word ``abstract'', since the class is a user interface component and SWT is short for Standard Widget Toolkit. As shown in the table, our model ranks it higher in the list than the baseline.

In the second case, the proposed algorithm assigns the word ``max'' a higher score than the baseline. The class ``java.io.inputstream'' is the superclass of all classes in Java representing an input stream of bytes, and the word ``max'' is used to describe the limitation of data being transferred.

Also in the third case, ``handler'' is more related to the Cascading Style Sheet class (``org.w3c.dom.css.cssvalue''). While in this case and the first case, CFA both ranks the textual feature ``abstract'' higher, it is not reasonable. Though the word may frequently appear in codes, but it is obviously not informative in our application,

Another interesting case is the last one. HMLCR puts a word ``exception'' on the top of our word list. As the word is very descriptive to the code feature ``java.io.ioexception'' and they both share the textual content ``exception'', the content based regularization clearly contributes to the performance.
\section{Related Work}\label{relatedwork}
To the best of our knowledge, there is no prior work that leverages codes themselves to enhance the software artifact retrieval task in the area of software engineering. The most similar work may be the task of automatic software traceability. Their methods focus on calculating the similarity between textual queries and the text contents of code. Another source of related work may come from cross-modal information retrieval, i.e., using queries in one media format (e.g., images) to retrieve objects in heterogeneous formats (e.g., video). In this section, we introduce some of the representative work in each domain.

Retrieving the code files by a query written in natural languages is first introduced in the area of software engineering and has been applied in a variety of tasks, such as development \cite{ref30} and maintenance \cite{ref24} of software, localization of concepts \cite{ref25} and features \cite{ref26}, tracing requirements back to source code \cite{ref27} and identify the corresponding code of a bug\cite{ref28,ref29}. Antoniol et al. \cite{ref1} first proposed to adopt Information Retrieval (IR) techniques to solve the problem. Naive Bayes and TF-IDF \cite{ref14} are used in their systems. More advanced approaches like Latent Semantic Indexing \cite{ref15} are incorporated to increase the accuracy \cite{ref4,ref5}. The researchers focus more on exploiting the domain specific rules to further improve their systems \cite{ref16,ref17} and better visualize the results \cite{ref5}. Some successful applications are summarized in \cite{ref2}. These methods are effective at solving the problem of source code understanding, retracing of requirements and maintaining IT systems. But besides the rule-based systems, of which the methods can hardly be generalized to similar problems, most existing work regards the code as general text. The functional information of code structures are ignored. In this work, the functional information is encoded in the code features.

Cross Modal information retrieval aims to build up a bridge between heterogeneous media by linking their features, which is very similar to the task of the proposed HMLCR model in this work, linking the code features and text features in a shared new space. In \cite{ref6}, they predefined a concept dictionary and proposed a model to map the visual features to the concepts. A similar work is proposed in \cite{ref7}, while in this work the concepts are replaced with the automatically produced word clusters. In \cite{ref18,ref19}, they seek to find the relatedness between the images and keywords. The aforementioned methods are designed for linking text and images, and the clustering of text and the clustering of image features are often separated: the clusters (latent topics) are first discovered and then the image features are mapped to the latent topics. Distance metric learning has also been used to transfer object recognition models to new domains, linking between visual features \cite{ref42}. A more general research topic is to find a heterogeneous distance metric between objects in different spaces \cite{ref20}. In \cite{ref8}, the joint graph regularization is incorporated. The heterogenous features are completely different in their task, while in our task, the code features and text features share useful information. The existing work on multimedia retrieval is not proper to be directly applied to the application, as the content information may be ignored.

\section{Conclusion and Future Work}\label{conclusion}
In this paper, based on the requirements of retrieval of software artifacts, we put forward a novel approach to calculate the similarity between code and text. We first formulated the problem and then proposed a novel feature extraction method for code programs. Two kinds of code features were proposed to exploit the functional and operational information of source codes, which are different from the textual features adopted by traditional methods. To calculate the similarity between codes and words, we introduced a novel heterogeneous distance metric learning approach to map the heterogeneous media into a unified space. Since the code features contain useful textual content, content-based regularization was further proposed to capture the content of source code. Content-based regularization distinguishes our work from the existing cross modal multimedia retrieval techniques. Data sets obtained from two open source projects were used in our experiments. Based on the real world data, such as bug reports, change logs and corresponding programs, experimental results validate the effectiveness of our proposed approach: the proposed code features and heterogeneous distance metric learning algorithms are helpful to improve the retrieval results, because both the proposed model and multimedia information retrieval methods outperform the traditional methods from the area of software engineering, which focuses on the textual part of programs; Content-based regularization proves to be useful, because it makes the cross modal information retrieval method compatible with our problem. A case study was also presented to further explain how the proposed framework worked.

In this work, HMLCR is used to bridge between code and text in the same project. But in real applications, identical code features, such as frequently occurred code patterns and class references, may exist in more than one projects. Thus, we also explore the relationship between different projects written in same language, and even projects written in different programming languages by leveraging the same code features and the co-occurrence of code and textual features. As a fundamental part of software development life cycle, software artifact retrieval is especially vital to large scale software project. So we will extend our algorithm to larger and more complex data sets in the future.

\section*{Acknowledgement}
We thank the support of the National Natural Science Foundation of China 91224006, the Strategic Priority Research Program of Chinese Academy of Sciences XDA06010202 and XDA05050601), ``12th Five Year'' Plan for Science \& Technology Support 2012BAK17B01 and 2013BAD15B02, the joint project by the Foshan and the Chinese Academy of Science under Grant No. 2012YS23, China National 973 program 2014CB340301.



%

\end{document}